\pdfoutput=1

\documentclass[11pt]{article}

\usepackage{eacl2023}

\usepackage{times}
\usepackage{latexsym}

\usepackage[T1]{fontenc}

\usepackage[utf8]{inputenc}

\usepackage{microtype}

\usepackage{inconsolata}

\usepackage{times}
\usepackage{latexsym}
\usepackage{natbib}
\usepackage{CJKutf8}
\usepackage{graphicx}
\usepackage{array}
\usepackage{svg}
\usepackage{multirow, booktabs}
\usepackage{amsmath,amssymb}
\usepackage{pgfplots}
\usetikzlibrary{patterns}

\newcommand{\lea}[1]{}
\newcommand{\leacom}[1]{}
\newcommand{\least}[1]{}
\newcommand{\oa}[1]{}
\newcommand{\oacom}[1]{}
\newcommand{\gabis}[1]{}
\newcommand{\gabisrep}[2]{}
\newcommand{\gabist}[1]{}
\newcommand{\gs}[1]{}
\newcommand{\ub}[1]{}
\newcommand{\ubst}[1]{}

\title{A Large-Scale Multilingual Study of Visual Constraints \\ on Linguistic Selection of Descriptions}

\author{Uri Berger$^{1,2}$, Lea Frermann$^2$, Gabriel Stanovsky$^1$, and Omri Abend$^1$ \\
$^1$School of Computer Science and Engineering, The Hebrew University of Jerusalem \\
$^2$School of Computing and Information Systems, University of Melbourne \\
\texttt{\{uri.berger2, gabriel.stanovsky, omri.abend\}@mail.huji.ac.il} \\
\texttt{lea.frermann@unimelb.edu.au}}

\begin{document}
\maketitle
\begin{abstract}
We present a large, multilingual study into how vision constrains linguistic choice, covering four languages and five linguistic properties, such as verb transitivity or use of numerals. 
We propose a novel method that leverages existing corpora of images with captions written by native speakers, and apply it to nine corpora, comprising 600k images and 3M captions. 
We study the relation between visual input and linguistic choices by training classifiers to predict the probability of expressing a property from raw images, and find evidence supporting the claim that linguistic properties are constrained by visual context across languages. 
We complement this investigation with a corpus study, taking the test case of numerals. Specifically, we use existing annotations (number or type of objects) to investigate the effect of different visual conditions on the use of numeral expressions in captions, and show that similar patterns emerge across languages.
Our methods and findings both confirm and extend existing research in the cognitive literature. We additionally discuss  possible applications for language generation.
We make our codebase publicly available.\footnote{\href{https://github.com/SLAB-NLP/visual_constraints_on_descriptions}{github.com/SLAB-NLP/visual\_constraints\_on\_descriptions}}
\end{abstract}

\section{Introduction}
In recent years, vision and language models have been shown to outperform models trained on a single modality in a variety of domains, such as language modeling~\cite{ororbia_et_al_2019}, document quality assessment~\cite{shen2020general}, and visual classification and segmentation~\cite{frome2013devise, radford_et_al_2021, berger2022clustering}.

\begin{figure}[tb]
    \centering
    \includegraphics[width=5cm]{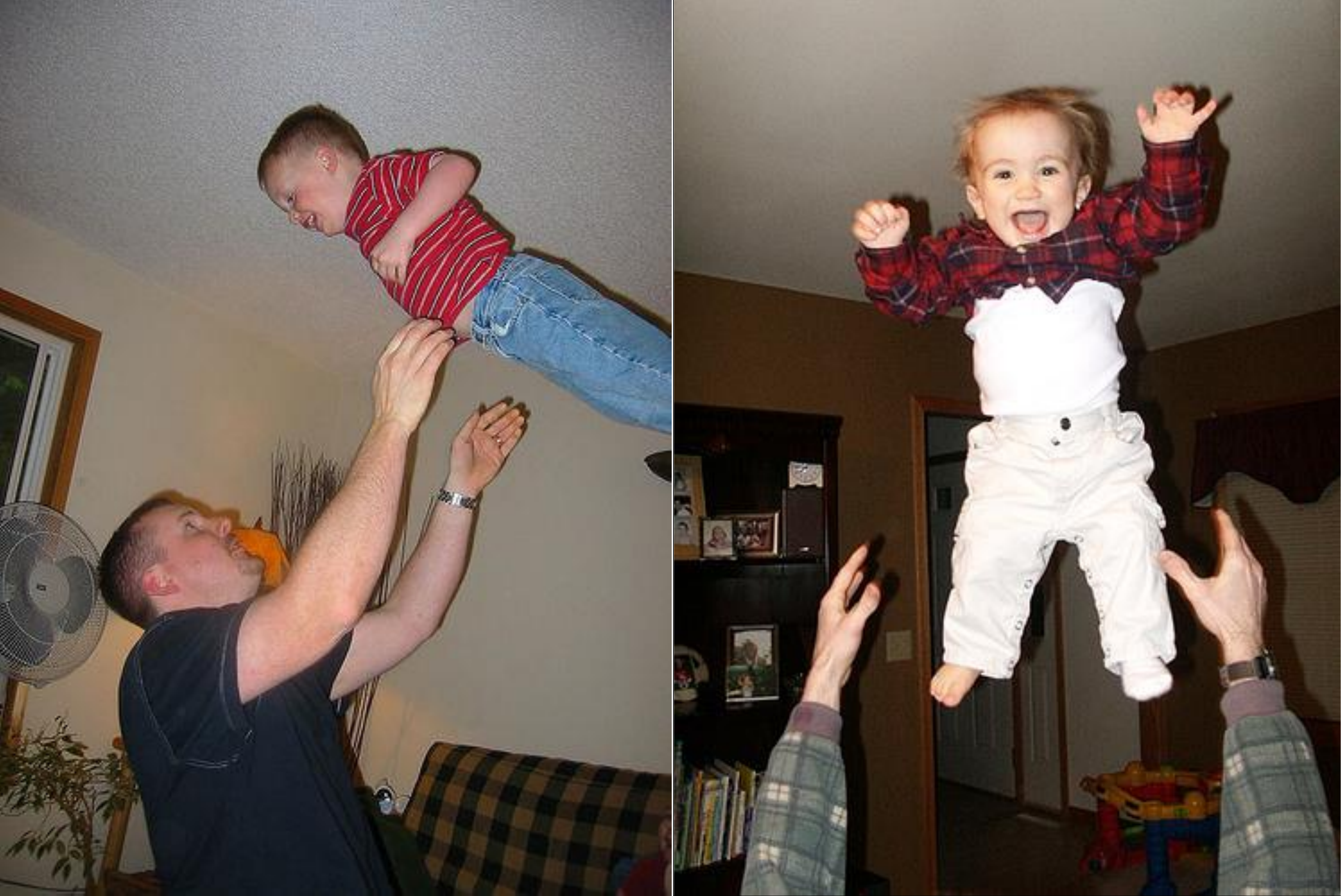}
    \caption{
    A demonstration of how visual cues in images may constrain linguistic choices in their captions. The image on the left in which the agent is visible is described using an active voice ``A man... is \emph{throwing} a child... in the air'', while in the right image the agent is not visible and the annotator chose a passive construction: ``A little boy... is \emph{thrown} in the air''. Images and captions taken from Flickr30k.
    }
    \label{fig:intro_image}
\end{figure}

While empirical evidence exists, including from the studies cited above, that each modality can benefit the other for task performance, less attention has been devoted to the broader question of {\it how} the two modalities influence and constrain one another. In this study, we focus on one aspect of this question: {\it How does vision constrain language?} We study the relation between the semantic content of an image and the language used to describe it.
As an example, consider Figure~\ref{fig:intro_image}. 
Captions of the right image, which crops the agent, use passive voice more frequently than those of the left \citep[taken from Flickr30k,][]{young2014image}.
We aim to study such trends by examining the influence of visual features of the image on the linguistic choices taken when describing it.

A number of psycholinguistic studies have aimed to answer this question by systematically varying visual conditions (such as image cropping) and analyzing verbal descriptions of the scenes by human participants for recurring differences \citep[e.g.,][]{chesney2015counts, rissman2019occluding}.
Although such controlled studies allow for precise measurement, the visual stimuli are synthetic (rather than depicting natural scenes), the manual annotation of descriptions limits the size of the dataset, and typically only one linguistic property is investigated in a single language.

We address this gap by proposing a scalable methodology that uses existing image-caption corpora in multiple languages.
We measure the correlation of visual features and linguistic properties of the caption by training visual classifiers to predict, for a given raw image, whether a linguistic property is expressed in its captions.
We compare the impact of different training sets (single vs. multiple languages) and different types of pre-training (none vs. object categories vs. visual vs.  textual pre-training objectives).
We use 9 large-scale image-caption datasets (overall 2.9M captions for 604K images), covering four languages (English, German, Chinese, Japanese), to study lexical properties (use of numerals and negation words) and structural properties (use of passive voice, transitivity of the main verb, choice of verbal vs. nominal constructions), which we automatically annotate in the captions. To the best of our knowledge, this is the first large-scale, multilingual study of the impact of visual input on linguistic choice.
We find evidence showing that the visual input imposes constraints on linguistic properties, and that such trends are detectable using the proposed methodology.

In a complementary corpus study, we link the prevalence of linguistic properties to existing, high-level visual annotations (number and type of objects), and find that these properties can be linked to the use of numeral expressions in similar patterns across languages, and in accordance with small-scale, highly-controlled psycholinguistic studies.

Our findings have both cognitive and computational implications. On the cognitive side, this study confirms findings from small-scale cognitive studies at scale: for naturalistic scenes, typologically diverse languages, and descriptions from thousands of native speakers. The magnitude of the data that can be studied with our method also allows the derivation of new insights, which can motivate additional controlled studies, making the proposed practice an effective exploration method.

On the computational side captioning models have been shown to generalize better when first predicting the syntactic structure of the generated caption~\cite{bugliarello2021role}. This research direction may benefit from the signal provided by our classifiers of linguistic properties.

\section{Background} \label{sec:background}
The notion that grammatical and lexical phenomena can be characterized semantically, at least in their prototypical instances, has a long tradition in linguistics \citep[among many others]{dixon1979ergativity,Goldberg1995ConstructionsAC,croft2012verbs}. For example, transitive clauses are often characterized as corresponding to actions instigated by volitional agents over passive objects, that are affected by the action. However, formally defining these semantic features in a non-linguistic way and showing empirically that the presence of these features indeed entails the presence of the corresponding linguistic feature, has proven to be methodologically challenging. One possible direction to address this is the use of images that implicitly define a type of non-linguistic semantics.
This section briefly reviews different approaches for studying visual constraints on our set of five phenomena from a cognitive and computational perspective (omitting phenomena that have not been previously covered).

\subsection{Cognitive Studies} \label{sec:cog_studies}

\textbf{Numerals.}
\emph{Subitizable} numbers are numbers that are rapidly and accurately visually counted by humans. Studies have shown that the threshold for subitizability is 4~\cite{kaufman1949discrimination, mandler1982subitizing}, with~\citet{barr2013generation} showing that humans tend to describe non-subitizable numbers using quantifiers (e.g.,  \emph{many}). In Section~\ref{sec:corpus_analysis} we confirm this result at scale.
\citet{chesney2015counts} asked participants to count objects in a given image, and found that participants were less likely to include objects located in frames (windows, mirrors or picture frames) in their count, suggesting that visual cues influence linguistic choices.

\textbf{Negation.}
Several studies have challenged the traditional view that images cannot express negation~\cite{worth1981studying, khemlani2012negation}. \citet{giora2009we} use visual negation markers (e.g., red cross road signs) to study neural processing of visual negation. \citet{oversteegen2014can} ask Dutch native speakers to describe images of objects missing integral parts (e.g., a woman without a mouth) and show that the descriptions are likely to contain a negation word.

\textbf{Passive voice.}
\citet{myachykov2012determinants} show that English native speakers have a stronger preference for using passive-voice when describing transitive events with visual cueing of their attention toward the agent (compared to the control condition). 

\textbf{Transitivity.}
\citet{rissman2019occluding} show that participants had a preference for intransitive descriptions of visual events (a person acting on an inanimate object) when the person was removed by cropping the image (whereas transitive descriptions were preferred in the base condition).

\subsection{Computational Studies}
\gabis{I prefer to put this in a related work section towards the end of the paper, let's get to our stuff as quickly as we can.}
Computational studies on how vision constrains language are rare. However, several studies examined various aspects of the linguistic properties studied in this work, typically focusing on individual properties and/or languages.

\textbf{Negation.}
A series of studies~\cite{van2016pragmatic, van2017cross}, investigated negation in Flickr30k image descriptions using a smaller set of negation words compared to our study, comparing the use of negation in English, German, and Dutch, and finding no significant differences.
\citet{dobreva2021investigating} show that the performance of vision and language models decreases when the text contains negation, but did not show that this decrease is caused by negation-related visual features. Text-only models also have difficulty processing negations~(e.g.,~\citet{ettinger2020bert}), and the drop in performance could be due to the text encoder alone.

The line of work most similar to this study train models to predict whether images from comics~\cite{sato2021visual} or real life~\cite{sato2021can} express negation, achieving chance-level results. In contrast to the current study, they used a single dataset, a single language (Japanese), and a single linguistic property (negation).

\textbf{Transitivity.}
\citet{nikolaus2019compositional} show that captioning models generalize better to unseen action -- object pairs when the action is transitive, hypothesizing that this improvement is due to the additional arguments (e.g. cake) that images describing transitive events (e.g. eating) contain.

\textbf{Verbal vs. nominal constructions.}
\citet{su2021dependency} study syntactic parsing and compare the Part-Of-Speech (POS) tag of the root of predicted and gold dependency trees of MSCOCO English captions, showing that the gold distribution is approximately 60:40 in favour of nouns, while models tend to never produce trees with a verb root.

\section{Approach} \label{sec:methodology}
We draw inspiration from the cognitive studies presented in Section~\ref{sec:cog_studies}. These studies carefully design visual scenes that differ only in terms of the visual feature of interest (e.g., visibility of the agent), ask
participants to describe the scenes, and compare the linguistic properties of the descriptions across different conditions. If a linguistic property is significantly more prevalent in one condition, it is assumed that this visual feature constrains that linguistic property.

While such setups allow careful control over the experimental design, they are also less ecologically valid in that they impose (1)~synthetic visual stimuli and (2)~limitations on the number and diversity of participants, phenomena and languages to include. We address both shortcomings by (1)~using large existing image caption datasets as a corpus of diverse language descriptions of naturalistic scenes, and (2)~annotating the captions automatically, yet accurately, with linguistic properties.

\begin{figure} [tb]
    \centering
    \includegraphics[width=7cm]{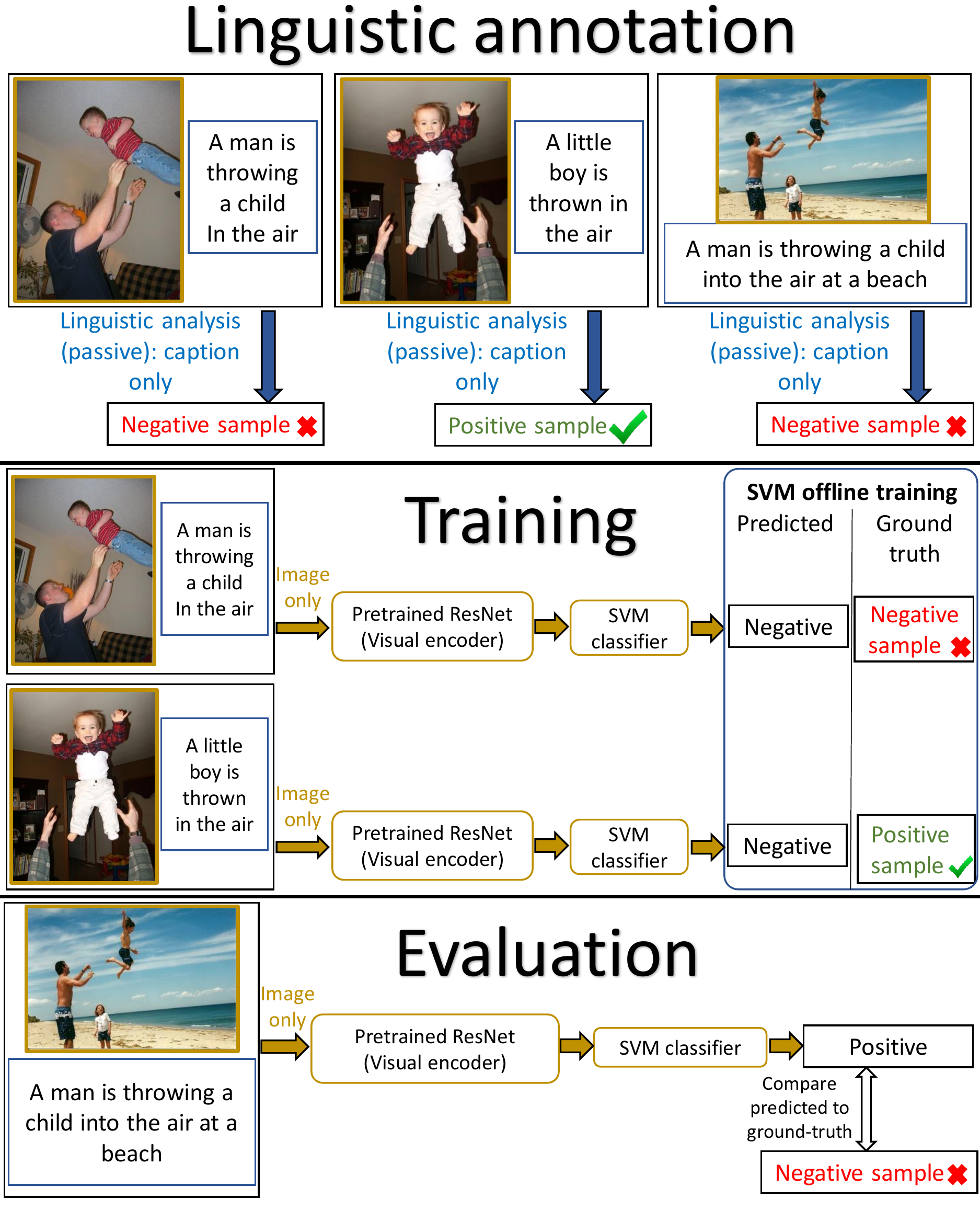}
    \caption{High-level depiction of our approach, demonstrating one linguistic property (passive voice). Top: Samples are annotated as expressing (positive) or not expressing (negative) the property. Middle: The SVM classifier is trained to predict whether an image expresses the property. Bottom: The classifier is evaluated; high classification accuracy indicates a strong relation between visual features and the linguistic property.}
    \label{fig:hl_desc}
\end{figure}

Using a large amount of data instead of a controlled experiment raises an issue.
Unlike in controlled cognitive studies, the sets of images we use are not arranged into `minimal pairs', which are identical except for a visual feature of interest. 
To overcome this limitation, we exploit the large amount of data available via the automatic annotation of linguistic properties. We train visual classifiers to predict if a linguistic property is expressed in image captions when only the image is provided. If the classifiers achieve high accuracy on a held out test set, it is an indication that the visual features are informative enough to predict the linguistic property.\footnote{However, if the accuracy is low, we cannot determine the cause; our modeling or data annotation assumptions may have led to this result, rather than the absence of a statistical relation.} Figure~\ref{fig:hl_desc} gives a high-level depiction of our approach.

To complement our analysis, we also conduct a corpus study. 
First, we use semantic annotations (object classes and bounding boxes) already available in existing datasets to group images by high-level properties and analyze the prevalence of linguistic properties in each group.
Second, we compare the linguistic properties of captions for the same image in different languages. If a property is salient in the captions of all languages for a given image, it is likely that its visual content constrains descriptions that use that property. We present a corpus analysis using both approaches in Section~\ref{sec:corpus_analysis}.

\section{Experimental Setup}
In this section we describe the languages~(\ref{sec:languages}), linguistic properties~(\ref{sec:prop_annotation}) and datasets~(\ref{sec:datasets}) used in our experiments.

\subsection{Languages} \label{sec:languages}

We study English (\textbf{En}), German (\textbf{De}), Chinese (\textbf{Zh}), and Japanese (\textbf{Jp}) for three main reasons. First, multiple language families are required to study language-agnostic constraints imposed by images. Second, all of these have large image-caption datasets with non-translated captions. Third, all of these have publicly available tools for annotation of some of the linguistic properties we study.

\subsection{Annotation of Linguistic Properties} \label{sec:prop_annotation}

Below, we describe the automatic annotation of occurrences of linguistic properties in captions.
All annotation methods were validated by asking in-house native speakers to verify a random sample of 100 (50 positive and 50 negative) instances per property and language. Across all languages and properties, accuracy exceeded 92\%, confirming that our automatic annotations are of high quality.

For Japanese we only study the use of numerals since we were not able to achieve accurate annotation for the other properties.

\textbf{Numerals (Num).}
We use Microsoft's Recognize-Text package\footnote{\href{https://github.com/microsoft/Recognizers-Text}{github.com/microsoft/Recognizers-Text}} to identify the use of numerals in all languages.
We ignore numerals with value of 1 for the following reasons: (1)
In German and Chinese, the same word can refer to the number \emph{one} or the determiner \emph{a};
(2) In Japanese, several non-numeral words contain the character for 1 
\begin{CJK*}{UTF8}{gbsn}
(一),
\end{CJK*}
confusing the recognizing algorithm.

\textbf{Negation words (Neg).}
We use the list of English negation words composed by~\citet{dobreva2021investigating}, and add the word \emph{nope}. We translate all words in the English list into the other languages, and verify the resulting lists with a native speaker.\footnote{All negation words are listed in Appendix~\ref{sec:app_prop_annotation}.}

\textbf{Verbal vs. nominal descriptions (Verb).}
We label captions with the root part-of-speech tag of their dependency tree, identified using Stanza's dependency parser~\cite{qi2020stanza}. We only consider captions with a single root which is a verb or a noun, filtering 0.8\% of the captions.
Note that we consider sentences where the root corresponds to the English verb \emph{to be} (\emph{sein} in German,
\begin{CJK*}{UTF8}{gbsn}
有
\end{CJK*}
in Chinese)
as noun roots, as no activity is described.

\textbf{Transitivity of main verb (Tran).}
We use Stanza's dependency parser and filter all captions with at least one of the following: (1) a non-verb root, (2) more or less than a single root, (3) the verb \emph{ be} (or its equivalents in languages other than English) as a root, filtering 47\% of the captions. After filtering, a caption is labeled as transitive if its root verb has a child labeled as a direct object, and intransitive otherwise.\footnote{In German and Chinese we automatically identify edge cases missed by the parser, see Appendix~\ref{sec:app_prop_annotation}.}

\textbf{Passive voice (Pass).}
We use the passive voice identifier tool for
English and German~\cite{ramm2017annotating}. For Chinese we search for the passive indicator
\begin{CJK*}{UTF8}{gbsn}
被,
\end{CJK*}
filtering cases where it is part of another word.\footnote{Words containing 
\begin{CJK*}{UTF8}{gbsn}
被
\end{CJK*}
are listed in Appendix~\ref{sec:app_prop_annotation}.}

\subsection{Datasets} \label{sec:datasets}

We use the following datasets: Pascal~\cite{rashtchian2010collecting}, MSCOCO~\cite{lin2014microsoft}, Flickr30k~\cite{young2014image}, Multi30k~\cite{elliott2016multi30k}, Flickr8kcn~\cite{li2016adding}, AIC-ICC~\cite{wu2017ai}, COCO-CN~\cite{li2019coco}, YJCaptions~\cite{miyazaki2016cross}, STAIR-captions~\cite{yoshikawa2017stair}. Table~\ref{tab:datasets} presents additional information.
We only use datasets with original captions generated by native speakers and avoid using datasets with captions translated from English.\footnote{See Appendix~\ref{sec:translated_captions_analysis} for a comparison of original and translated captions.} In addition to captions, MSCOCO and Flickr30k contain object classes and bounding box annotations.
A description of the data collection process for each dataset is provided in Appendix~\ref{sec:dataset_collect}.

\begin{table}[!t]
\centering
\scalebox{0.75}{
\begin{tabular}{cccc}
\toprule
Lan &
Name &
Based on &
Size (im,cap)
\\
\midrule
En &
Pascal &
& 1k, 5k \\
& MSCOCO &
& 123k, 616k \\
& Flickr30k &
& 31k, 158k \\
& \bf{Overall} &
& 156k, 779k \\
\midrule
De &
Multi30k & Flickr30k
& 31k, 155k \\
& \bf{Overall} &
& 31k, 155k \\
\midrule
Zh & Flickr8kCN & Flickr8k
& 8k, 40k \\
& AIC-ICC &
& 240k, 1.2M \\
& COCO-CN & MSCOCO
& 20k, 22k \\
& \bf{Overall} &
& 268k, 1.262M \\
\midrule
Jp & YJCaptions & MSCOCO
& 26k, 131k \\
& STAIR-captions & MSCOCO
& 123k, 616k \\
& \bf{Overall} &
& 149k, 747k \\
\bottomrule
\end{tabular}}
\caption{
The list of datasets used in this study.
For non-English datasets based on images from an English dataset, the \emph{Based on} column indicates the name of the original dataset.
The \emph{Size} column indicates the number of images and the number of captions in the dataset.
}
\label{tab:datasets}
\end{table}

\textbf{Property expression probability.}
Given an image $I$, a set of captions $c_1, ..., c_{N_I}$, and an annotation function $f_p(c_i)$ mapping caption $c_i$ to $1$ if it expresses property $p$, and to $0$ otherwise, we define the probability that image $I$ expresses $p$ as

\begin{equation*}
P_p(I) = \frac{\sum_{i=1}^{N_I} f_p(c_i)}{N_I}
\end{equation*}

Given a language $\mathcal{L}$ we denote with  $P_{p,\mathcal{L}}$ the same computation with the set of captions filtered to include only captions in $\mathcal{L}$.
Given a set of images $S$ we denote the expected probability of expressing property $p$ across all images in $S$ as

\begin{equation*}
\mathbb{E}_{I \in S}[P_p(I)] = \frac{\sum_{I \in S} P_p(I)}{|S|}
\end{equation*}

Table~\ref{tab:prevalence} presents the expected probability of each property $p$ occurring in captions in language $\mathcal{L}$, $\mathbb{E}_{I \in S_\mathcal{L}}[P_{p,\mathcal{L}}(I)]$, where $S_\mathcal{L}$ is the set of all images in all datasets of language $\mathcal{L}$.

\begin{table}
\centering
\scalebox{0.75}{
\begin{tabular}{c>{\centering\arraybackslash}p{1cm}>{\centering\arraybackslash}p{1.2cm}>{\centering\arraybackslash}p{1.2cm}>{\centering\arraybackslash}p{1cm}>{\centering\arraybackslash}p{1cm}} 
\toprule
 & Num & Neg & Pass & Tran & Verb
\\
\midrule
En & 0.13 (156k) & 0.0046 (156k) & 0.076 (148k) & 0.35 (137k) & 0.41 (156k)
\\
\midrule
De & 0.19 (31k) & 0.0024 (31k) & 0.012 (31k) & 0.25 (31k) & 0.78 (31k)
\\
\midrule
Zh & 0.34 (268k) & 0.0002 (268k) & 0.002 (268k) & 0.48 (245k) & 0.60 (268k)
\\
\midrule
Jp & 0.13 (123k) & -- & -- & -- & --
\\
\bottomrule
\end{tabular}}
\caption{
Expected probability of images expressing each property, in each of the 4 languages. Number of images are in parentheses.
}
\label{tab:prevalence}
\end{table}

\section{Experiments} \label{sec:experiments}
We now describe our experiments and analyses. In Section~\ref{sec:classifiers} we train visual classifiers to predict linguistic properties, Section~\ref{sec:corpus_analysis} presents a complementary corpus analysis, and Section~\ref{sec:insights} presents additional insights that may lead to future research.

\subsection{Predicting Properties from Images} \label{sec:classifiers}

We study the task of predicting, given an image, whether human annotators will use a particular linguistic property when describing it. The input is a raw image and the output is binary, indicating whether the descriptions express the property.

\textbf{Models.}
Our model consists of a visual encoder~\citep[ResNet50,][]{he_et_al_2016} to embed the raw image, followed by a set of binary SVM classifiers, one per linguistic property.\footnote{We also experimented with neural classifiers, but SVM performed significantly better: see Appendix \ref{sec:neural_classifier} for details.} We investigate four different pre-training methods with varying levels of supervision from different modalities.

First, we randomly initialize the visual encoder (no pre-training; \textbf{None}), avoiding unwanted bias through pre-training with human annotated information. Using a random encoder renders the task for the classifier more difficult, and the classifier might perform poorly even if linguistic properties are highly correlated with visual features, so we consider \textbf{None} as a lower bound.

To equip our model with some prior visual knowledge, we use MoCo~\cite{he2020momentum}, a self-supervised pre-training method based only on visual signals (\textbf{MoCo}). MoCo creates multiple manipulated versions of an image and trains the encoder to predict if two manipulated images correspond to the same original.

We also include ImageNet~\cite{deng_et_al_2009} pre-training (\textbf{IN}). The visual encoder is first trained to classify images in the ImageNet dataset, and then the classification head is discarded. Although semantic information is provided in ImageNet pre-training through class-labels, no textual input is provided which {\it describes} the visual scene.

Finally, we use \textbf{CLIP}~\cite{radford_et_al_2021} pre-training. CLIP is a multimodal self-supervised model, trained to project images and corresponding captions to similar vectors in a joint space. We use CLIP's visual encoder, discarding the text encoder. This method is pre-trained with explicit textual input, and hence its predictions will be skewed by the prior probability of linguistic properties in general language, obscuring the correlation with image features. In terms of raw performance, we consider CLIP as an upper bound.

We study two settings: monolingual (all images from datasets in a single language) and multilingual (all images from all datasets).
In each setting, for each linguistic property $p$, we compute the probability of all relevant images to express $p$ and binarize the data by using the median probability value as a threshold above which the image is considered to express $p$. Finally, we create a balanced dataset\footnote{Although balancing the test set is usually considered a bad practice, in this study we only study image-text correlation and our classifiers would not be used for classifying new samples in the future.} of images
that express $p$ and those that do not. We evaluate our models using 5-fold cross-validation.
Table~\ref{tab:generated_datasets} shows the statistics of the generated datasets (note that the size of the datasets is smaller than in Table~\ref{tab:prevalence} because the data was balanced using down sampling). Implementation details are in Appendix~\ref{sec:model_details}.

\begin{table}
\centering
\begin{small}
\begin{tabular}{rrrrrr}
\toprule
& En & De & Zh &
Jp &
Mul
\\ \midrule
Numerals & 88k & 21k & 224k & 
86k &
337k
\\
Passive & 47k & 2k & 2k &
-- &
50k
\\
Negation & 6k & 0.7k & 0.3k & 
-- &
7k
\\
Transitivity & 128k & 29k & 223k & 
-- &
339k
\\
Verb Root & 150k & 20k & 198k & 
-- &
333k
\\ \bottomrule
\end{tabular}
\end{small}
\caption{
Number of images used in the experiments, for all properties and languages (Mul: Multilingual).
}
\label{tab:generated_datasets}
\end{table}

\paragraph{Multilingual results.}
Results are presented in Table~\ref{tab:multilingual_results}. First, we observe that except for the model without pre-training, all models predict all properties above chance levels, supporting the hypothesis that linguistic properties are constrained by visual context.
Second, results for the two non-textual pre-training methods (MoCo, IN) were significantly higher than the lower bound (None) and lower than the upper bound (CLIP) in all properties.
Finally, numerals seem easiest to predict, which concurs with our corpus analysis where we find that mentions of numerals were easiest to link to visual properties (Section~\ref{sec:corpus_analysis}).

\paragraph{Monolingual results.}
We applied MoCo, the best performing method without human annotated pre-training, individually to each language (Table~\ref{tab:monolingual_results}). Note that model performance does not always correlate with training data size (Table~\ref{tab:generated_datasets}): in English, the verb root dataset was the largest but the classifier achieved the lowest accuracy; and prediction accuracy was high for passive voice in Chinese despite a small dataset.
Across all languages, use of numerals was predicted most reliably.

\begin{table}
\centering
\scalebox{0.68}{
\begin{tabular}{cccccc}
\toprule
 & Num & Pass & Neg & Tran & Verb
\\
\midrule
None & \textbf{60.5 $\pm 0.9$} & 52.7 $\pm 0.3$ & 51.0 $\pm 1.6$ & 54.3 $\pm 0.5$ & 54.2 $\pm 0.3$
\\
\midrule
MoCo & \textbf{76.4 $\pm 0.2$} & 66.2 $\pm 0.4$ & 62.6 $\pm 1.2$ & 64.7 $\pm 0.3$ & 63.1 $\pm 0.2$
\\
IN & \textbf{74.6 $\pm 0.4$} & 65.9 $\pm 0.5$ & 62.4 $\pm 1.9$ & 64.5 $\pm 0.1$ & 62.6 $\pm 0.2$
\\
\midrule
CLIP & \textbf{81.4 $\pm 0.2$} & 68.2 $\pm 0.2$ & 65.3 $\pm 1.5$ & 68.7 $\pm 0.3$ & 65.4 $\pm 0.1$
\\
\bottomrule
\end{tabular}}
\caption{
Multilingual classification 5-fold cross-validation accuracy on all linguistic properties and pre-training methods. In all configurations, chance level is 50. IN: ImageNet. Numerals is the highest scoring property (in bold).
}
\label{tab:multilingual_results}
\end{table}

\begin{table}
\centering
\scalebox{0.7}{
\begin{tabular}{cccccc} 
\bottomrule
 & Num & Pass & Neg & Tran & Verb
\\
\midrule
En & \textbf{68.3 $\pm 0.3$} & 66.8 $\pm 0.7$ & 62.5 $\pm 0.8$ & 64.6 $\pm 0.3$ & 58.8 $\pm 0.1$
\\
De & \textbf{69.5 $\pm 0.6$} & 58.5 $\pm 3.1$ & 51.5 $\pm 4.3$ & 62.0 $\pm 0.7$ & 57.8 $\pm 0.8$
\\
Zh & \textbf{80.6 $\pm 0.2$} & 70.9 $\pm 2.6$ & 55.4 $\pm 4.3$ & 65.8 $\pm 0.3$ & 67.3 $\pm 0.2$
\\
Jp & \textbf{67.4 $\pm 0.3$} & -- & -- & -- & --
\\
\bottomrule
\end{tabular}}
\caption{
Monolingual classification 5-fold cross-validation accuracy on all linguistic properties and languages, using the MoCo pre-training method. In all configurations, chance level is 50. In all languages, the use of numerals was predicted most accurately (in bold).
}
\label{tab:monolingual_results}
\end{table}

\subsection{Corpus Analysis} \label{sec:corpus_analysis}

In this section we show that large image captioning corpora not only allow us to build predictive models to test hypotheses about the constraints of visual properties on language, but also support large-scale corpus studies. Our goal is to correlate image properties (e.g., the type or number of objects in an image) with linguistic choice (e.g., the use of numerals). The ground truth image properties are typically unavailable, but we can use additional information in MSCOCO and Flickr30k as proxies. In particular, we use the fact that the corpora are multilingually aligned (each image contains captions in different languages, all generated by native speakers) and they contain additional annotations (class labels and bounding boxes). 

We take the expression of numerals as a test case, since it was the one most accurately predicted in Section \ref{sec:classifiers}. We emphasize, however, that the approach generalizes to other properties as well.


Although both MSCOCO and Flickr30k contain object classes and bounding box annotations, MSCOCO's granularity is much higher (80 classes compared to 10 classes), so we only use MSCOCO's class and bounding box annotations in our analysis.
German is excluded from the class and bounding box analysis as there is no German version of MSCOCO with original captions.

\paragraph{Images containing animals are most likely to be described using numerals across languages.}
For each MSCOCO class $c$, we find the set $S_c$ of all images instantiating that class and compute $\mathbb{E}_{I \in S_c}[P_{\text{num}}(I)]$. We note that the expected $P_{\text{num}}(I)$ of some classes might be lower simply because they are more likely to occur in singles, and avoid this bias by filtering out images with a single instantiation of $c$ from $S_c$.\footnote{No classes were completely filtered out; only two classes (\emph{toaster}, \emph{hair-drier}) were left with less than 80 images.}

\begin{figure} [tb]
    \centering
    \includegraphics[width=8cm]{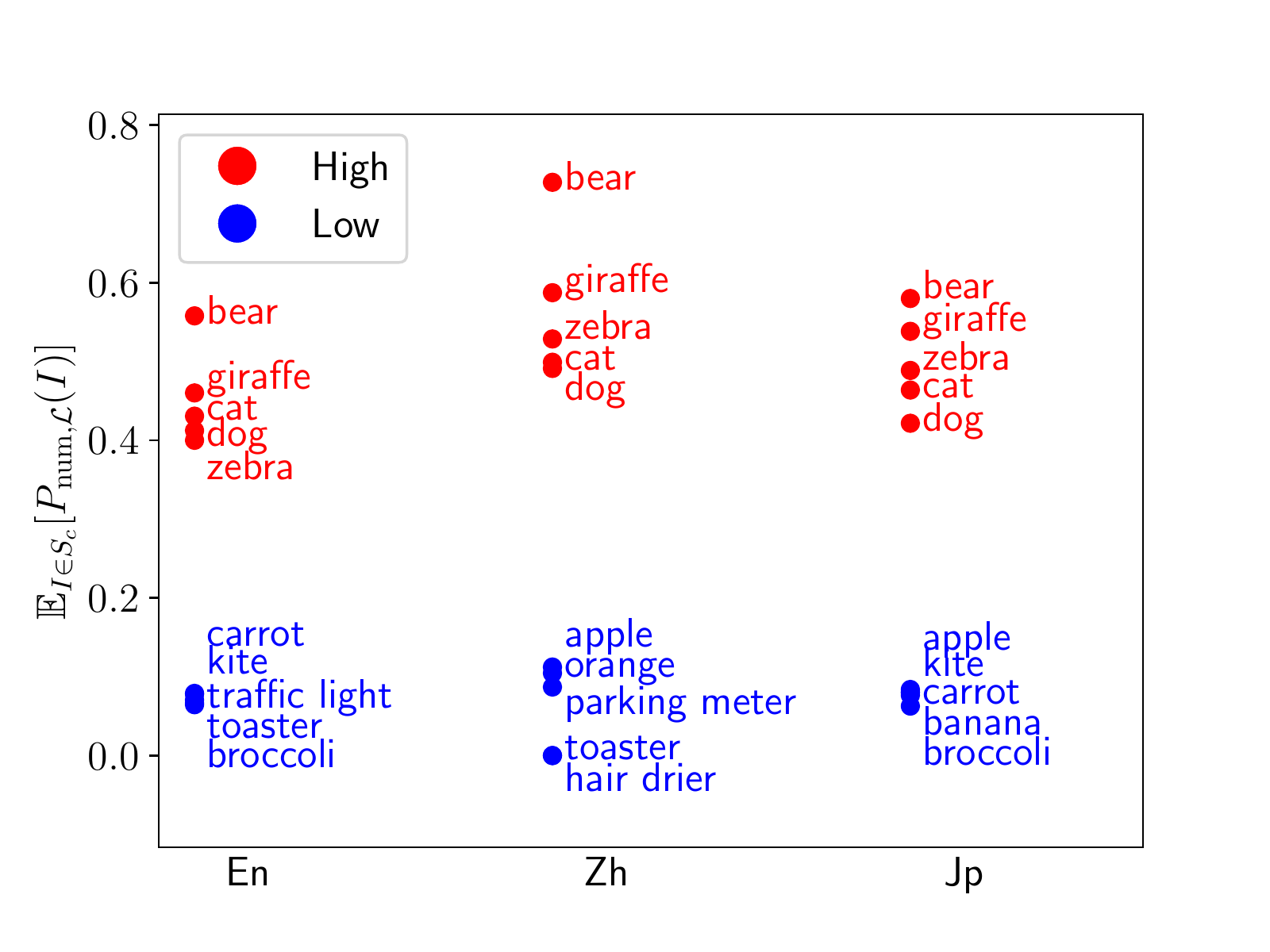}
    \caption{MSCOCO classes with highest and lowest expected numeral expression probability $\mathbb{E}_{I \in S_c} [P_{\text{num}, \mathcal{L}}(I)]$, for
    $\mathcal{L} \in \{ \text{En}, \text{Zh}, \text{Jp} \}$.
    The probability of classes of animals is high in all languages.}
    \label{fig:num_coco_classes}
\end{figure}

Figure~\ref{fig:num_coco_classes} shows the 5 classes with the highest and lowest $\mathbb{E}_{I \in S_c}[P_{\text{num}}(I)]$ for each language. In all languages, images depicting animals are most likely to be described with numerals.

\begin{figure}[tb]
    \centering
    \includegraphics[width=8cm]{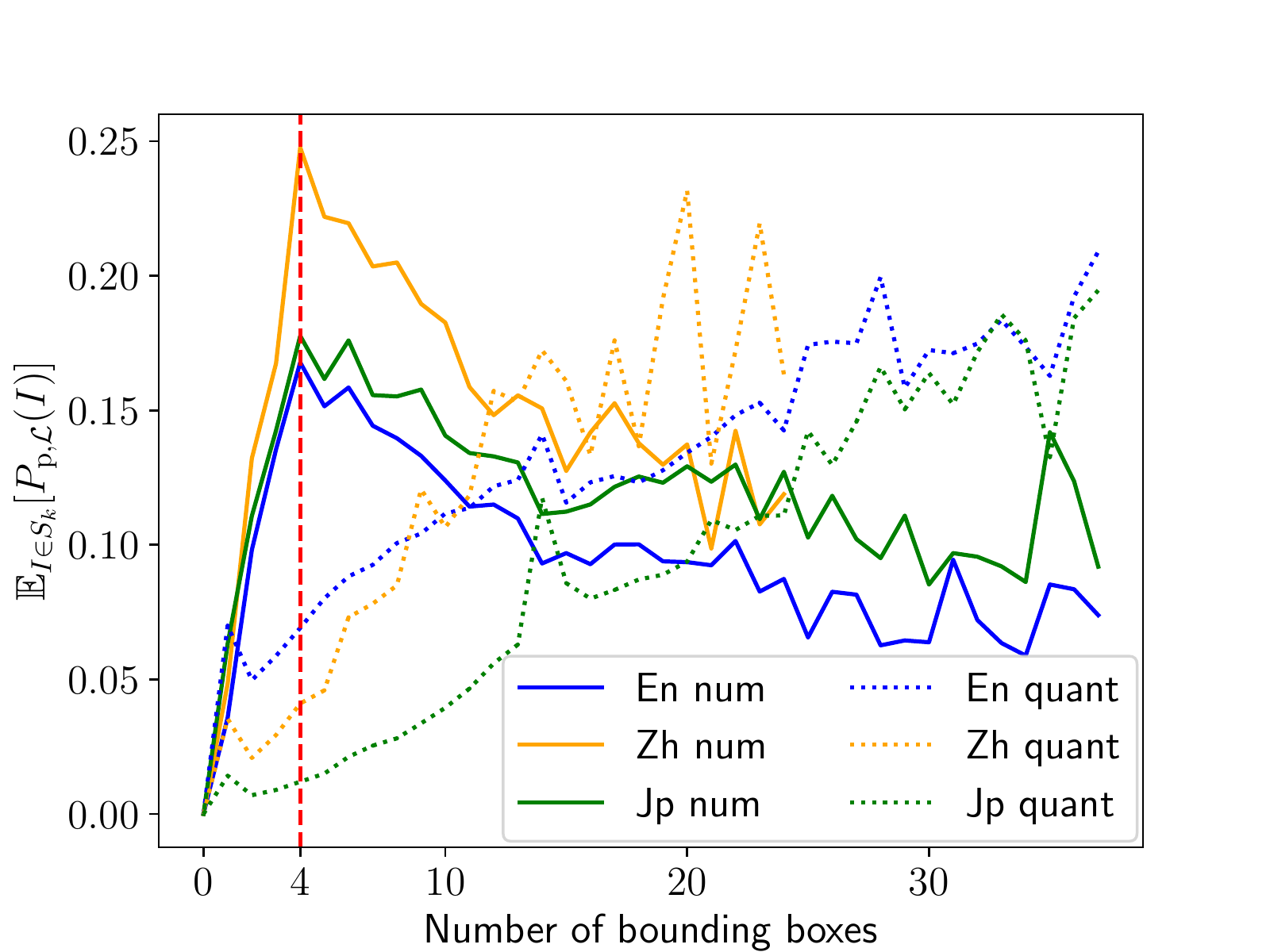}
    \caption{Expected probability of expressing the use of numerals and quantifiers $\mathbb{E}_{I \in S_k} [P_{p, \mathcal{L}}(I)]$ as a function of the number of bounding boxes in MSCOCO, for
    $\mathcal{L} \in \{ \text{En}, \text{Zh}, \text{Jp} \}$ and $p \in \{ \text{num}, \text{quant} \}$.
    All $k$ with $|S_k| < 100$ were removed from the plot. Red line: subitizability threshold. In all languages, the probability increases up to 4 objects (consistent with cognitive studies) and then decreases. Quantifiers expression probability increases steadily.}
    \label{fig:num_coco_bboxes}
\end{figure}

\paragraph{Our findings corroborate cognitive findings, placing the human subitizability threshold at 4.}
We use MSCOCO bounding boxes annotation to investigate whether the use of numerals in image descriptions reflects the subitizability threshold (see Section~\ref{sec:cog_studies}).
For each integer $k$, we find the set $S_k$ of all images with $k$ labeled bounding boxes, and compute $\mathbb{E}_{I \in S_k}[P_{\text{num}}(I)]$. We also label captions with quantifiers (e.g., \emph{some, a few}\footnote{The full lists are in Appendix~\ref{sec:app_prop_annotation}.}) and compute $\mathbb{E}_{I \in S_k}[P_{\text{quant}}(I)]$.
Figure~\ref{fig:num_coco_bboxes} shows the results, for all $k$ where $|S_k| \geq 100$.
In all languages, $\mathbb{E}_{I \in S_k}[P_{\text{num}}(I)]$ initially increases with a clear peak at 4, while quantifiers expression steadily increases.

\paragraph{Captions of the same image in different languages tend to agree on numerals usage.}
We use the multilingual datasets Flickr30k (En, De, Zh) and MSCOCO
(En, Zh, Jp), identify a list of images with captions in all respective languages $\{ I_k \}_{k=1}^{N}$, and compute the list of probabilities of numerals expression for each image $L_{\mathcal{L}} = \{ P_{\text{num}, \mathcal{L}}(I_k) \}_{k=1}^{N}$, in each language $\mathcal{L}$. Next, we compute the Pearson correlation coefficient of $L_{\mathcal{L}_1},L_{\mathcal{L}_2}$ for each pair of languages $\mathcal{L}_1,\mathcal{L}_2$. The results are shown in Table~\ref{tab:num_agreement}. The correlation is {\it high} ($>0.5$;  \citet{cohen2013statistical}) across all languages and datasets.

\begin{table}
\centering
\begin{small}
\begin{tabular}{lll|lll}
\toprule
\multicolumn{3}{c}{\textbf{Flickr30k (7k)}} & \multicolumn{3}{c}{\textbf{MSCOCO (13k)}}\\
Zh/De & Zh/En & De/En & Zh/En & Zh/Jp & Jp/En\\ \midrule
0.75 & 0.72 & 0.87 & 0.59 & 0.52 & 0.68\\\bottomrule
\end{tabular}
\end{small}
\caption{Pearson correlation of the probability of use of numerals across pairs of languages in Flickr30 and MSCOCO. Number of images in parentheses.}
\label{tab:num_agreement}
\end{table}


\subsection{Additional Insights} \label{sec:insights}

The proposed methodology can also be used as an exploration method for further cognitive research. In this section, we present findings obtained by manually investigating extreme cases of property expression. This is an exploratory analysis, presenting preliminary findings that may lead to future research in a more controlled setting.

\begin{figure}
    \centering
    \includegraphics[width=6cm]{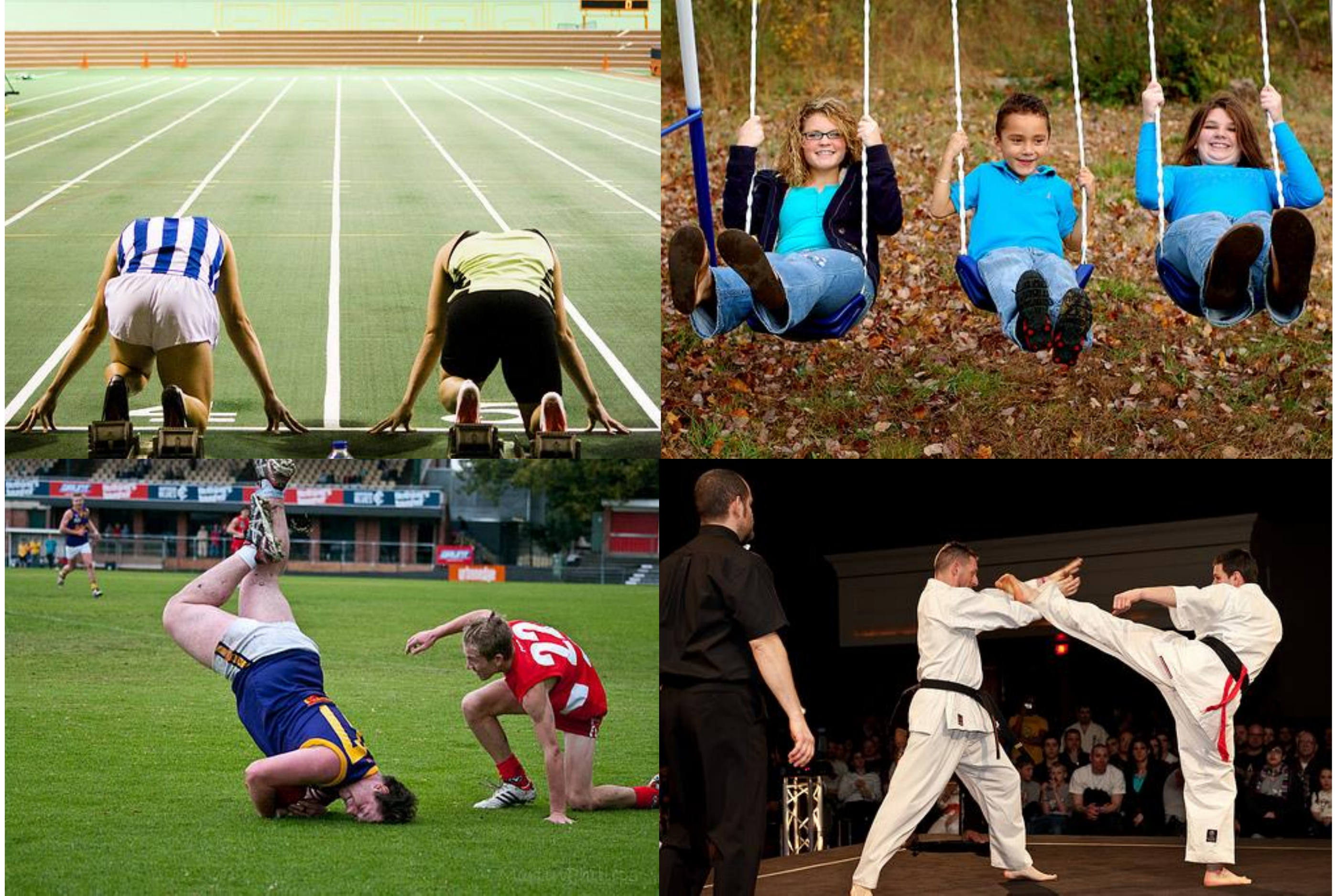}
    \caption{Top: images described using numerals in all languages. Bottom: images described without numerals. Images taken from Flickr30k.}
    \label{fig:cons_high_images}
\end{figure}

\textbf{Use of numeral expressions.}
We manually inspect all images that use numerals in all captions across all languages in Flickr30k (N=105). The top images in Figure~\ref{fig:cons_high_images} are representative examples. All images depict multiple participants taking similar roles and positioned in a regular pattern (e.g., all the children in the upper right image in Figure~\ref{fig:cons_high_images} are swinging and facing the camera). 
The bottom of Figure~\ref{fig:cons_high_images} shows comparable images, which were never described using numerals. Here, participants appear in different poses and roles. 
We hypothesize that people count more easily and accurately when objects are arranged in a regular pattern, compared to a random formation~\cite{burgess1983precision}. 

We also present differences in the use of numerals across languages.
We analyze images for which at least two captions use numerals with the same numeral value in each language, but different values across languages (N=46).
We find two main reasons for cross-language inconsistencies: First, different languages tend to either count all participants in a single group or split them into smaller groups based on gender, role, or age.\footnote{Examples for all partition types are in Appendix~\ref{sec:app_visual_examples}.}
These differences may be due to different annotation guidelines or different cultural backgrounds of the annotators.

\begin{figure} [tb]
\centering
\begin{small}
\begin{tabular}{p{5cm}l}
{\bf En:} Basketball player wearing a
white, number 23 jersey jumps up
with the ball while guarded by
number 13 on the opposite team & \multirow{5}{*}{\includegraphics[width=0.12\textwidth,clip,trim=23cm 2cm 0 0]{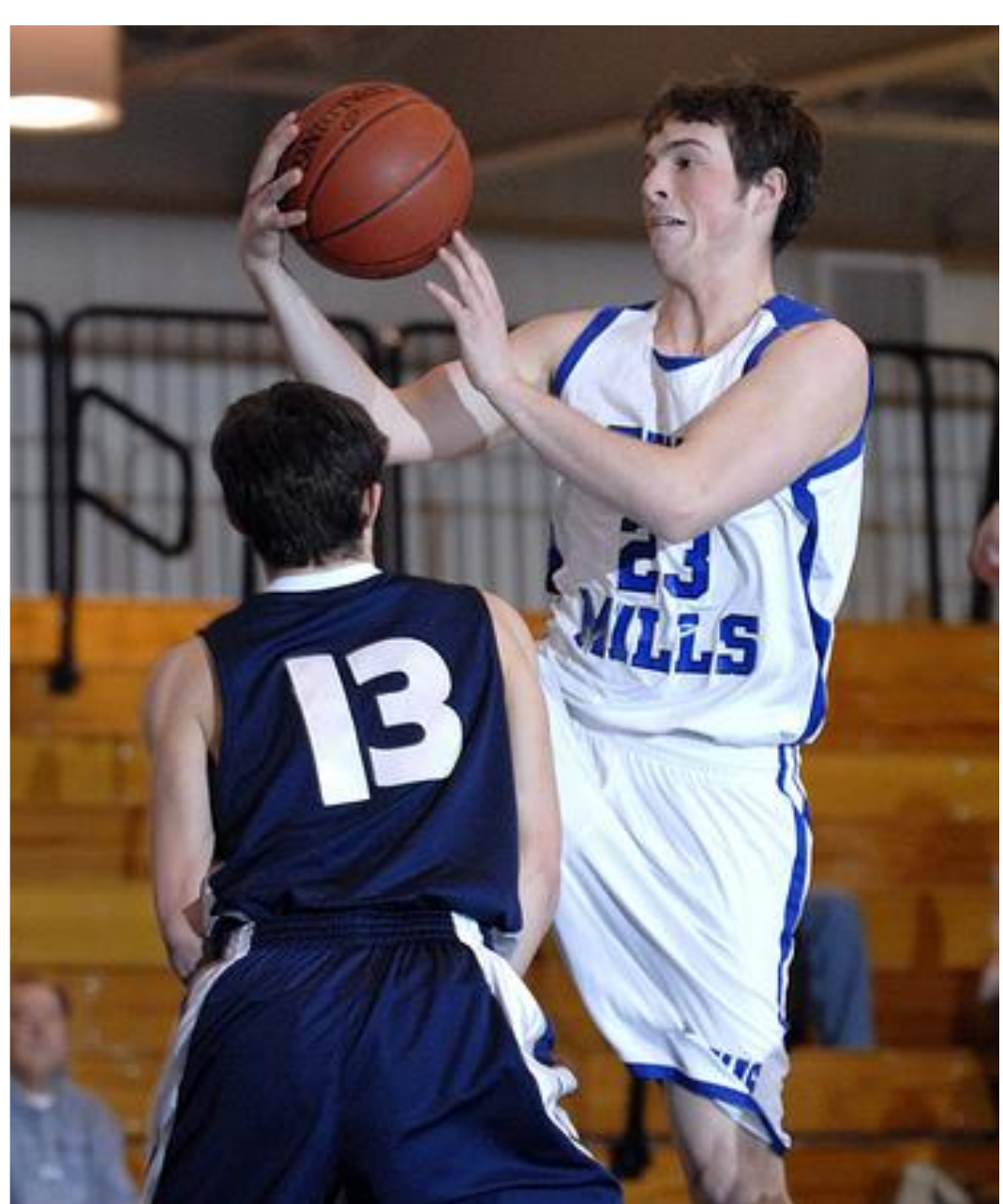}}\\
{\bf De: } Zwei Männer spielen Basketball\\
{\bf Zh:} \begin{CJK*}{UTF8}{gbsn}
有两个男人正在打篮球
\end{CJK*}
\end{tabular}
\end{small}
    \caption{An image of a basketball game. The English captions are highly detailed, while both the German and Chinese caption translates to {\it Two men are playing basketball}. Image taken from Flickr30k, captions taken from Flickr30k (En), Mutli30k (De), Flickr8kcn (Zh).}
    \label{fig:sports_dis}
\end{figure}

Second, the multilingual datasets were originally created for English captioning, making the selected images highly related to English and especially North American culture.\footnote{This is a well known problem in multimodal datasets, previously discussed by~\citet{liu2021visually}.}
For example, in the sports domain,
the datasets contain mainly images of Basketball and Baseball, popular sports in the United States. While English annotators use a detailed description, commonly mentioning the players' shirt number, German and Chinese descriptions are mostly short and count the number of players in the image (Figure~\ref{fig:sports_dis}).

\textbf{Passive.}
\label{ssec:corpus_passive}
We notice that in images with high probability for using passive voice, the patient is commonly located at the center of the scene either by the pose of the camera or the borders of the image. We hypothesize that this visual feature is correlated with the use of passive voice. The right image of Figure~\ref{fig:intro_image} shows one example. More examples are in Appendix~\ref{sec:app_visual_examples}.


\subsection{Discussion}

Our experiments suggest that various linguistic properties are predictable from visual context, most notably in the case of the use of numeral expressions.
Our classifiers were able to predict the presence of numerals in captions with high accuracy. Correspondingly, our corpus analysis provides evidence that the type and number of objects in the image constrain the use of numerals. Both results hold across different languages, and present high agreement between languages in the selection of images that are described with numerals. This lends support to the hypothesis that visual context constrains the use of numerals across a variety of languages from different families, and that such trends can be studied using the proposed methodology.

A surprising result is that without pretraining of the visual encoder (\textbf{None}), above chance-level performance can be obtained, most notably for the numerals property. A randomly initialized visual encoder applies a random dimensionality reduction to the input image, and the fact that the SVM classifier was able to learn to predict the presence of numerals in the captions of images at above chance level following this random transformation supports the hypothesis that this property correlates with visual features.



\section{Conclusion}
The synergistic relation between vision and language has been shown in the cognitive literature and leveraged in computational models, but {\it how} the two modalities inform each other has not been sufficiently studied at scale. We present a large scale study of the correlation of visual properties with linguistic phenomena, using naturalistic images described by a large crowd of native speakers of four languages. 

In addition to confirming results of previous cognitive studies, we present new findings, e.g., the effect of object type on the use of numerals in visual scene description and the cross-lingual correlation of the use of numerals.
Considering the effort needed to execute a controlled study, our proposed method can be used as an effective exploration technique for finding hypotheses for future controlled studies.
In addition, our framework is general, and extends naturally to more languages and properties.

Beyond the cognitive contribution, our work can inform NLP models.
Recent work suggests that  in captioning models, training the model to predict a structured representation of the caption (e.g., based on POS prediction) before the text improves compositional generalization~\cite{bugliarello2021role}. In future work, we will study the utility of predicting our proposed linguistic properties for improving captioning models.

\section*{Limitations} \label{sec:limitations}
We acknowledge several limitations of our suggested methodology. First, confounding factors may have affected our results, e.g., the difference in wording of the annotation guidelines for the original image-caption dataset could have a significant impact on the linguistic properties of the descriptions. In cognitive research, there is a well-known trade-off and ongoing discussion on the merits of highly-controlled, yet often oversimplified settings and the larger-scale, yet typically confounded, studies. The ``reproducibility crisis'' has highlighted that controlled studies are often difficult to reproduce, and initiated a discussion about the (complementary) utility of large-scale experiments which are typically more realistic. We propose such a method in the context of language/vision research, which can complement small-scale cognitive studies by considering natural scenes, while covering several languages and linguistic phenomena. We present empirical results that support the validity of the methodology, in the sense that it often accords with established findings from the literature, as well as small scale qualitative analysis, that suggests trends for future work. We emphasize the importance of both paradigms, which should coexist and complement one another.

Second, with the exception of AIC-ICC, all image collections for all languages are based on original English image-caption datasets and hence are Anglocentric in their selection of concepts. The impact of such bias on NLP research has recently been discussed~\cite{liu2021visually}.
We hope to extend the analysis with additional culture- and language-specific datasets in the future.

Finally, we do not distinguish between differences in linguistic properties that are due to annotators' focus choices (i.e., the selection of what details in the image to describe) and those that are due to linguistic choices. The prevalence of linguistic properties could be influenced by the content that the annotator chose to describe (e.g., some annotators describe the background in addition to the main object(s), and others do not). This is a challenging and important line of future work, but outside the scope of this study.

\section*{Acknowledgements}
We would like to thank the anonymous reviewers for their helpful comments and feedback.
We would also like to thank Rotem Dror, Sharon Goldwater and Grzegorz Chrupała for consulting, and the native speakers that consulted and validated our annotation tool: Assaf Porat, Kozue Watanabe, Arie Cattan, and Yilin Geng.
This work was supported in part by the Israel Science Foundation (grant no. 2424/21), the Israeli Ministry of Science and Technology (grant no. 2336), and by the HUJI-UoM joint PhD program.

\section*{Ethics Statement}
We use publicly available resources in our experiments, in accordance with their license agreements. The datasets are fully anonymized and do not contain personal information about the caption annotators or any information that could reveal the identity of the photographed subjects.


\bibliography{bib}

\begin{thebibliography}{47}
\expandafter\ifx\csname natexlab\endcsname\relax\def\natexlab#1{#1}\fi

\bibitem[{Barr et~al.(2013)Barr, van Deemter, and
  Fern{\'a}ndez}]{barr2013generation}
Dale Barr, Kees van Deemter, and Raquel Fern{\'a}ndez. 2013.
\newblock \href {https://aclanthology.org/W13-2120} {Generation of quantified
  referring expressions: Evidence from experimental data}.
\newblock In \emph{Proceedings of the 14th {E}uropean Workshop on Natural
  Language Generation}, pages 157--161, Sofia, Bulgaria. Association for
  Computational Linguistics.

\bibitem[{Berger et~al.(2022)Berger, Stanovsky, Abend, and
  Frermann}]{berger2022clustering}
Uri Berger, Gabriel Stanovsky, Omri Abend, and Lea Frermann. 2022.
\newblock \href {https://aclanthology.org/2022.naacl-main.280} {A computational
  acquisition model for multimodal word categorization}.
\newblock In \emph{Proceedings of the 2022 Conference of the North American
  Chapter of the Association for Computational Linguistics}.

\bibitem[{Bugliarello and Elliott(2021)}]{bugliarello2021role}
Emanuele Bugliarello and Desmond Elliott. 2021.
\newblock \href {https://doi.org/10.18653/v1/2021.eacl-main.48} {The role of
  syntactic planning in compositional image captioning}.
\newblock In \emph{Proceedings of the 16th Conference of the European Chapter
  of the Association for Computational Linguistics: Main Volume}, pages
  593--607, Online. Association for Computational Linguistics.

\bibitem[{Burgess and Barlow(1983)}]{burgess1983precision}
A~Burgess and HB~Barlow. 1983.
\newblock The precision of numerosity discrimination in arrays of random dots.
\newblock \emph{Vision Research}, 23(8):811--820.

\bibitem[{Chesney and Gelman(2015)}]{chesney2015counts}
Dana~L Chesney and Rochel Gelman. 2015.
\newblock What counts? visual and verbal cues interact to influence what is
  considered a countable thing.
\newblock \emph{Memory \& cognition}, 43(5):798--810.

\bibitem[{Cohen(2013)}]{cohen2013statistical}
Jacob Cohen. 2013.
\newblock \emph{Statistical power analysis for the behavioral sciences}.
\newblock Routledge.

\bibitem[{Croft(2012)}]{croft2012verbs}
William Croft. 2012.
\newblock \emph{Verbs: Aspect and causal structure}.
\newblock OUP Oxford.

\bibitem[{Deng et~al.(2009)Deng, Dong, Socher, Li, Li, and
  Li}]{deng_et_al_2009}
Jia Deng, Wei Dong, Richard Socher, Li{-}Jia Li, Kai Li, and Fei{-}Fei Li.
  2009.
\newblock \href {https://doi.org/10.1109/CVPR.2009.5206848} {Imagenet: {A}
  large-scale hierarchical image database}.
\newblock In \emph{2009 {IEEE} Computer Society Conference on Computer Vision
  and Pattern Recognition {(CVPR} 2009), 20-25 June 2009, Miami, Florida,
  {USA}}, pages 248--255. {IEEE} Computer Society.

\bibitem[{Dixon(1979)}]{dixon1979ergativity}
R.~M.~W. Dixon. 1979.
\newblock Ergativity.
\newblock \emph{Language}, 55:59–138.

\bibitem[{Dobreva and Keller(2021)}]{dobreva2021investigating}
Radina Dobreva and Frank Keller. 2021.
\newblock \href {https://doi.org/10.18653/v1/2021.blackboxnlp-1.27}
  {Investigating negation in pre-trained vision-and-language models}.
\newblock In \emph{Proceedings of the Fourth BlackboxNLP Workshop on Analyzing
  and Interpreting Neural Networks for NLP}, pages 350--362, Punta Cana,
  Dominican Republic. Association for Computational Linguistics.

\bibitem[{Elliott et~al.(2016)Elliott, Frank, Sima{'}an, and
  Specia}]{elliott2016multi30k}
Desmond Elliott, Stella Frank, Khalil Sima{'}an, and Lucia Specia. 2016.
\newblock \href {https://doi.org/10.18653/v1/W16-3210} {{M}ulti30{K}:
  Multilingual {E}nglish-{G}erman image descriptions}.
\newblock In \emph{Proceedings of the 5th Workshop on Vision and Language},
  pages 70--74, Berlin, Germany. Association for Computational Linguistics.

\bibitem[{Ettinger(2020)}]{ettinger2020bert}
Allyson Ettinger. 2020.
\newblock \href {https://doi.org/10.1162/tacl_a_00298} {What {BERT} is not:
  Lessons from a new suite of psycholinguistic diagnostics for language
  models}.
\newblock \emph{Transactions of the Association for Computational Linguistics},
  8:34--48.

\bibitem[{Everingham et~al.(2008)Everingham, Zisserman, Williams, Van~Gool,
  Allan, Bishop, Chapelle, Dalal, Deselaers, Dork{\'o}
  et~al.}]{everingham2008pascal}
Mark Everingham, Andrew Zisserman, Christopher~KI Williams, Luc Van~Gool, Moray
  Allan, Christopher~M Bishop, Olivier Chapelle, Navneet Dalal, Thomas
  Deselaers, Gyuri Dork{\'o}, et~al. 2008.
\newblock The pascal visual object classes challenge 2007 (voc2007) results.

\bibitem[{Frome et~al.(2013)Frome, Corrado, Shlens, Bengio, Dean, Ranzato, and
  Mikolov}]{frome2013devise}
Andrea Frome, Gregory~S. Corrado, Jonathon Shlens, Samy Bengio, Jeffrey Dean,
  Marc'Aurelio Ranzato, and Tom{\'{a}}s Mikolov. 2013.
\newblock \href
  {https://proceedings.neurips.cc/paper/2013/hash/7cce53cf90577442771720a370c3c723-Abstract.html}
  {Devise: {A} deep visual-semantic embedding model}.
\newblock In \emph{Advances in Neural Information Processing Systems 26: 27th
  Annual Conference on Neural Information Processing Systems 2013. Proceedings
  of a meeting held December 5-8, 2013, Lake Tahoe, Nevada, United States},
  pages 2121--2129.

\bibitem[{Giora et~al.(2009)Giora, Heruti, Metuki, and Fein}]{giora2009we}
Rachel Giora, Vered Heruti, Nili Metuki, and Ofer Fein. 2009.
\newblock “when we say no we mean no”: Interpreting negation in vision and
  language.
\newblock \emph{Journal of Pragmatics}, 41(11):2222--2239.

\bibitem[{Goldberg(1995)}]{Goldberg1995ConstructionsAC}
Adele~E. Goldberg. 1995.
\newblock Constructions: A construction grammar approach to argument structure.
\newblock University of Chicago Press.

\bibitem[{Gupta et~al.(2020)Gupta, Lenka, Ekbal, and
  Bhattacharyya}]{gupta2020unified}
Deepak Gupta, Pabitra Lenka, Asif Ekbal, and Pushpak Bhattacharyya. 2020.
\newblock \href {https://aclanthology.org/2020.aacl-main.90} {A unified
  framework for multilingual and code-mixed visual question answering}.
\newblock In \emph{Proceedings of the 1st Conference of the Asia-Pacific
  Chapter of the Association for Computational Linguistics and the 10th
  International Joint Conference on Natural Language Processing}, pages
  900--913, Suzhou, China. Association for Computational Linguistics.

\bibitem[{He et~al.(2020)He, Fan, Wu, Xie, and Girshick}]{he2020momentum}
Kaiming He, Haoqi Fan, Yuxin Wu, Saining Xie, and Ross~B. Girshick. 2020.
\newblock \href {https://doi.org/10.1109/CVPR42600.2020.00975} {Momentum
  contrast for unsupervised visual representation learning}.
\newblock In \emph{2020 {IEEE/CVF} Conference on Computer Vision and Pattern
  Recognition, {CVPR} 2020, Seattle, WA, USA, June 13-19, 2020}, pages
  9726--9735. {IEEE}.

\bibitem[{He et~al.(2016)He, Zhang, Ren, and Sun}]{he_et_al_2016}
Kaiming He, Xiangyu Zhang, Shaoqing Ren, and Jian Sun. 2016.
\newblock \href {https://doi.org/10.1109/CVPR.2016.90} {Deep residual learning
  for image recognition}.
\newblock In \emph{2016 {IEEE} Conference on Computer Vision and Pattern
  Recognition, {CVPR} 2016, Las Vegas, NV, USA, June 27-30, 2016}, pages
  770--778. {IEEE} Computer Society.

\bibitem[{Hitschler et~al.(2016)Hitschler, Schamoni, and
  Riezler}]{hitschler2016multimodal}
Julian Hitschler, Shigehiko Schamoni, and Stefan Riezler. 2016.
\newblock \href {https://doi.org/10.18653/v1/P16-1227} {Multimodal pivots for
  image caption translation}.
\newblock In \emph{Proceedings of the 54th Annual Meeting of the Association
  for Computational Linguistics (Volume 1: Long Papers)}, pages 2399--2409,
  Berlin, Germany. Association for Computational Linguistics.

\bibitem[{Kaufman et~al.(1949)Kaufman, Lord, Reese, and
  Volkmann}]{kaufman1949discrimination}
Edna~L Kaufman, Miles~W Lord, Thomas~Whelan Reese, and John Volkmann. 1949.
\newblock The discrimination of visual number.
\newblock \emph{The American journal of psychology}, 62(4):498--525.

\bibitem[{Khemlani et~al.(2012)Khemlani, Orenes, and
  Johnson-Laird}]{khemlani2012negation}
Sangeet Khemlani, Isabel Orenes, and Philip~N Johnson-Laird. 2012.
\newblock Negation: A theory of its meaning, representation, and use.
\newblock \emph{Journal of Cognitive Psychology}, 24(5):541--559.

\bibitem[{Li et~al.(2016)Li, Lan, Dong, and Liu}]{li2016adding}
Xirong Li, Weiyu Lan, Jianfeng Dong, and Hailong Liu. 2016.
\newblock Adding chinese captions to images.
\newblock In \emph{Proceedings of the 2016 ACM on international conference on
  multimedia retrieval}, pages 271--275.

\bibitem[{Li et~al.(2019)Li, Xu, Wang, Lan, Jia, Yang, and Xu}]{li2019coco}
Xirong Li, Chaoxi Xu, Xiaoxu Wang, Weiyu Lan, Zhengxiong Jia, Gang Yang, and
  Jieping Xu. 2019.
\newblock Coco-cn for cross-lingual image tagging, captioning, and retrieval.
\newblock \emph{IEEE Transactions on Multimedia}, 21(9):2347--2360.

\bibitem[{Lin et~al.(2014)Lin, Maire, Belongie, Hays, Perona, Ramanan,
  Doll{\'a}r, and Zitnick}]{lin2014microsoft}
Tsung-Yi Lin, Michael Maire, Serge Belongie, James Hays, Pietro Perona, Deva
  Ramanan, Piotr Doll{\'a}r, and C~Lawrence Zitnick. 2014.
\newblock Microsoft coco: Common objects in context.
\newblock In \emph{European conference on computer vision}, pages 740--755.
  Springer.

\bibitem[{Liu et~al.(2021)Liu, Bugliarello, Ponti, Reddy, Collier, and
  Elliott}]{liu2021visually}
Fangyu Liu, Emanuele Bugliarello, Edoardo~Maria Ponti, Siva Reddy, Nigel
  Collier, and Desmond Elliott. 2021.
\newblock \href {https://doi.org/10.18653/v1/2021.emnlp-main.818} {Visually
  grounded reasoning across languages and cultures}.
\newblock In \emph{Proceedings of the 2021 Conference on Empirical Methods in
  Natural Language Processing}, pages 10467--10485, Online and Punta Cana,
  Dominican Republic. Association for Computational Linguistics.

\bibitem[{Mandler and Shebo(1982)}]{mandler1982subitizing}
George Mandler and Billie~J Shebo. 1982.
\newblock Subitizing: an analysis of its component processes.
\newblock \emph{Journal of experimental psychology: general}, 111(1):1.

\bibitem[{Miyazaki and Shimizu(2016)}]{miyazaki2016cross}
Takashi Miyazaki and Nobuyuki Shimizu. 2016.
\newblock \href {https://doi.org/10.18653/v1/P16-1168} {Cross-lingual image
  caption generation}.
\newblock In \emph{Proceedings of the 54th Annual Meeting of the Association
  for Computational Linguistics (Volume 1: Long Papers)}, pages 1780--1790,
  Berlin, Germany. Association for Computational Linguistics.

\bibitem[{Myachykov et~al.(2012)Myachykov, Garrod, and
  Scheepers}]{myachykov2012determinants}
Andriy Myachykov, Simon Garrod, and Christoph Scheepers. 2012.
\newblock Determinants of structural choice in visually situated sentence
  production.
\newblock \emph{Acta psychologica}, 141(3):304--315.

\bibitem[{Nikolaus et~al.(2019)Nikolaus, Abdou, Lamm, Aralikatte, and
  Elliott}]{nikolaus2019compositional}
Mitja Nikolaus, Mostafa Abdou, Matthew Lamm, Rahul Aralikatte, and Desmond
  Elliott. 2019.
\newblock \href {https://doi.org/10.18653/v1/K19-1009} {Compositional
  generalization in image captioning}.
\newblock In \emph{Proceedings of the 23rd Conference on Computational Natural
  Language Learning (CoNLL)}, pages 87--98, Hong Kong, China. Association for
  Computational Linguistics.

\bibitem[{Ororbia et~al.(2019)Ororbia, Mali, Kelly, and
  Reitter}]{ororbia_et_al_2019}
Alexander Ororbia, Ankur Mali, Matthew Kelly, and David Reitter. 2019.
\newblock \href {https://doi.org/10.18653/v1/P19-1506} {Like a baby: Visually
  situated neural language acquisition}.
\newblock In \emph{Proceedings of the 57th Annual Meeting of the Association
  for Computational Linguistics}, pages 5127--5136, Florence, Italy.
  Association for Computational Linguistics.

\bibitem[{Oversteegen and Schilperoord(2014)}]{oversteegen2014can}
Eleonore Oversteegen and Joost Schilperoord. 2014.
\newblock Can pictures say no or not? negation and denial in the visual mode.
\newblock \emph{Journal of pragmatics}, 67:89--106.

\bibitem[{Qi et~al.(2020)Qi, Zhang, Zhang, Bolton, and Manning}]{qi2020stanza}
Peng Qi, Yuhao Zhang, Yuhui Zhang, Jason Bolton, and Christopher~D. Manning.
  2020.
\newblock \href {https://doi.org/10.18653/v1/2020.acl-demos.14} {{S}tanza: A
  python natural language processing toolkit for many human languages}.
\newblock In \emph{Proceedings of the 58th Annual Meeting of the Association
  for Computational Linguistics: System Demonstrations}, pages 101--108,
  Online. Association for Computational Linguistics.

\bibitem[{Radford et~al.(2021)Radford, Kim, Hallacy, Ramesh, Goh, Agarwal,
  Sastry, Askell, Mishkin, Clark, Krueger, and Sutskever}]{radford_et_al_2021}
Alec Radford, Jong~Wook Kim, Chris Hallacy, Aditya Ramesh, Gabriel Goh,
  Sandhini Agarwal, Girish Sastry, Amanda Askell, Pamela Mishkin, Jack Clark,
  Gretchen Krueger, and Ilya Sutskever. 2021.
\newblock \href {http://proceedings.mlr.press/v139/radford21a.html} {Learning
  transferable visual models from natural language supervision}.
\newblock In \emph{Proceedings of the 38th International Conference on Machine
  Learning, {ICML} 2021, 18-24 July 2021, Virtual Event}, volume 139 of
  \emph{Proceedings of Machine Learning Research}, pages 8748--8763. {PMLR}.

\bibitem[{Ramm et~al.(2017)Ramm, Lo{\'a}iciga, Friedrich, and
  Fraser}]{ramm2017annotating}
Anita Ramm, Sharid Lo{\'a}iciga, Annemarie Friedrich, and Alexander Fraser.
  2017.
\newblock \href {https://aclanthology.org/P17-4001} {Annotating tense, mood and
  voice for {E}nglish, {F}rench and {G}erman}.
\newblock In \emph{Proceedings of {ACL} 2017, System Demonstrations}, pages
  1--6, Vancouver, Canada. Association for Computational Linguistics.

\bibitem[{Rashtchian et~al.(2010)Rashtchian, Young, Hodosh, and
  Hockenmaier}]{rashtchian2010collecting}
Cyrus Rashtchian, Peter Young, Micah Hodosh, and Julia Hockenmaier. 2010.
\newblock \href {https://aclanthology.org/W10-0721} {Collecting image
  annotations using {A}mazon{'}s {M}echanical {T}urk}.
\newblock In \emph{Proceedings of the {NAACL} {HLT} 2010 Workshop on Creating
  Speech and Language Data with {A}mazon{'}s Mechanical Turk}, pages 139--147,
  Los Angeles. Association for Computational Linguistics.

\bibitem[{Rissman et~al.(2019)Rissman, Woodward, and
  Goldin-Meadow}]{rissman2019occluding}
Lilia Rissman, Amanda Woodward, and Susan Goldin-Meadow. 2019.
\newblock Occluding the face diminishes the conceptual accessibility of an
  animate agent.
\newblock \emph{Language, cognition and neuroscience}, 34(3):273--288.

\bibitem[{Sato and Mineshima(2021)}]{sato2021can}
Yuri Sato and Koji Mineshima. 2021.
\newblock Can humans and machines classify photographs as depicting negation?
\newblock In \emph{International Conference on Theory and Application of
  Diagrams}, pages 348--352. Springer.

\bibitem[{Sato et~al.(2021)Sato, Mineshima, and Ueda}]{sato2021visual}
Yuri Sato, Koji Mineshima, and Kazuhiro Ueda. 2021.
\newblock \href {https://arxiv.org/abs/2105.10131} {Visual representation of
  negation: Real world data analysis on comic image design}.
\newblock \emph{ArXiv preprint}, abs/2105.10131.

\bibitem[{Shen et~al.(2020)Shen, Salehi, Qi, and Baldwin}]{shen2020general}
Aili Shen, Bahar Salehi, Jianzhong Qi, and Timothy Baldwin. 2020.
\newblock A general approach to multimodal document quality assessment.
\newblock \emph{Journal of Artificial Intelligence Research}, 68:607--632.

\bibitem[{Su et~al.(2021)Su, Rijhwani, Zhu, He, Wang, Bisk, and
  Neubig}]{su2021dependency}
Ruisi Su, Shruti Rijhwani, Hao Zhu, Junxian He, Xinyu Wang, Yonatan Bisk, and
  Graham Neubig. 2021.
\newblock \href {https://doi.org/10.18653/v1/2021.conll-1.2} {Dependency
  induction through the lens of visual perception}.
\newblock In \emph{Proceedings of the 25th Conference on Computational Natural
  Language Learning}, pages 17--26, Online. Association for Computational
  Linguistics.

\bibitem[{van Miltenburg et~al.(2017)van Miltenburg, Elliott, and
  Vossen}]{van2017cross}
Emiel van Miltenburg, Desmond Elliott, and Piek Vossen. 2017.
\newblock \href {https://doi.org/10.18653/v1/W17-3503} {Cross-linguistic
  differences and similarities in image descriptions}.
\newblock In \emph{Proceedings of the 10th International Conference on Natural
  Language Generation}, pages 21--30, Santiago de Compostela, Spain.
  Association for Computational Linguistics.

\bibitem[{van Miltenburg et~al.(2016)van Miltenburg, Morante, and
  Elliott}]{van2016pragmatic}
Emiel van Miltenburg, Roser Morante, and Desmond Elliott. 2016.
\newblock \href {https://doi.org/10.18653/v1/W16-3207} {Pragmatic factors in
  image description: The case of negations}.
\newblock In \emph{Proceedings of the 5th Workshop on Vision and Language},
  pages 54--59, Berlin, Germany. Association for Computational Linguistics.

\bibitem[{Worth(1981)}]{worth1981studying}
Sol Worth. 1981.
\newblock Studying visual communication.
\newblock In \emph{Studying Visual Communication}. University of Pennsylvania
  Press.

\bibitem[{Wu et~al.(2017)Wu, Zheng, Zhao, Li, Yan, Liang, Wang, Zhou, Lin, Fu
  et~al.}]{wu2017ai}
Jiahong Wu, He~Zheng, Bo~Zhao, Yixin Li, Baoming Yan, Rui Liang, Wenjia Wang,
  Shipei Zhou, Guosen Lin, Yanwei Fu, et~al. 2017.
\newblock \href {https://arxiv.org/abs/1711.06475} {Ai challenger: A
  large-scale dataset for going deeper in image understanding}.
\newblock \emph{ArXiv preprint}, abs/1711.06475.

\bibitem[{Yoshikawa et~al.(2017)Yoshikawa, Shigeto, and
  Takeuchi}]{yoshikawa2017stair}
Yuya Yoshikawa, Yutaro Shigeto, and Akikazu Takeuchi. 2017.
\newblock \href {https://doi.org/10.18653/v1/P17-2066} {{STAIR} captions:
  Constructing a large-scale {J}apanese image caption dataset}.
\newblock In \emph{Proceedings of the 55th Annual Meeting of the Association
  for Computational Linguistics (Volume 2: Short Papers)}, pages 417--421,
  Vancouver, Canada. Association for Computational Linguistics.

\bibitem[{Young et~al.(2014)Young, Lai, Hodosh, and
  Hockenmaier}]{young2014image}
Peter Young, Alice Lai, Micah Hodosh, and Julia Hockenmaier. 2014.
\newblock \href {https://doi.org/10.1162/tacl_a_00166} {From image descriptions
  to visual denotations: New similarity metrics for semantic inference over
  event descriptions}.
\newblock \emph{Transactions of the Association for Computational Linguistics},
  2:67--78.

\end{thebibliography}
\bibliographystyle{acl_natbib}

\appendix

\section{Implementation Details}
\label{sec:impl_details}

\subsection{Linguistic Properties Annotation} \label{sec:app_prop_annotation}

\paragraph{Use of numerals}
In the bounding boxes experiment in section \ref{sec:corpus_analysis} we search for quantifiers. Following are the lists of quantifiers we search for in each language. English: \emph{some, a lot of, many, lots of, a few, several, a number of}. Chinese:
\begin{CJK*}{UTF8}{gbsn}
些, 多.
\end{CJK*}
Japanese:
\begin{CJK*}{UTF8}{gbsn}
多くの, たくさん, いくつか.
\end{CJK*}

\paragraph{Use of negation words}
Following are the lists of negation words for each language.

English: \emph{not, isn't, aren't, doesn't, don't, can't, cannot, shouldn't, wont, wouldn't, no, none, nobody, nothing, nowhere, neither, nor, never, without, nope}.

German: \emph{nicht, kein, nie, niemals, niemand, nirgendwo, nirgendwohin, nirgends, weder, ohne, nein, nichts, nee}. We lemmatize the words in the sentence before searching in this list.

Chinese:
\begin{CJK*}{UTF8}{gbsn}
不, 不是, 不能, 不可以, 没, 没有, 没什么, 从不, 并不, 从没有, 并没有, 无人, 无处, 无, 别, 绝不.
\end{CJK*}
We use the Jieba tokenizer\footnote{\href{https://github.com/fxsjy/jieba}{github.com/fxsjy/jieba}}. We also identify cases where one of the words above is part of a longer non-negation word and filter those cases. Following is the list of non-negation words:
\begin{CJK*}{UTF8}{gbsn}
别着, 不小心, 不一样.
\end{CJK*}

\paragraph{Use of passive verbs}
In Chinese we search for the passive indicator
\begin{CJK*}{UTF8}{gbsn}
被,
\end{CJK*}
filtering cases where it is part of the
\begin{CJK*}{UTF8}{gbsn}
被子
\end{CJK*}
word (meaning quilt), a common word in the AIC-ICC dataset.

\paragraph{Transitivity}
In German and Chinese we identify several important edge cases in which the Stanza parser is consistently incorrect, which we fix manually. All edge cases were verified by native speakers.

In German we identify sentences containing a node which is a child of the root and labeled with the \emph{PTKVZ} POS tag, and label these as intransitive.

In Chinese we identify sentences where (1) the lemma of the root word ends with the preposition token
\begin{CJK*}{UTF8}{gbsn}
在;
\end{CJK*}
(2) the lemma of the word following the root word is
\begin{CJK*}{UTF8}{gbsn}
在;
\end{CJK*}
or (3) the lemma of the word following the root word starts with the preposition token
\begin{CJK*}{UTF8}{gbsn}
向,
\end{CJK*}
and label these as intransitive.

\subsection{Model Details} \label{sec:model_details}

\subsubsection{SVM Classifier}

We use the SVC model from the sklearn Python package with the RBF kernel and default hyper-parameters.

\subsubsection{Neural Classifier} \label{sec:neural_classifier}

We use a feed-forward neural network with 1 or 2 hidden layers, with different activation functions (ReLU, Sigmoid, Tanh). In all configurations, the SVM classifier performed better.

\subsubsection{Pre-trained Backbone Models}

For MoCo and CLIP we use the models provided in the officially published code. For ImageNet pre-training we use the pre-trained model provided by the PyTorch package. In all cases, model contains 25.6M parameters.

\subsection{Training}

Training with the largest training set (the transitivity multilingual setting, see table \ref{tab:generated_datasets}) took 30 hours on a single GM204GL GPU.

\section{Dataset Collection Details} \label{sec:dataset_collection}
\label{sec:dataset_collect}

Following is a brief description of the process of data collection for each of the datasets.

\textbf{Pascal Sentences} \cite{rashtchian2010collecting} contains the set of images from the PASCAL Visual Object Classes Challenge \cite{everingham2008pascal} with captions generated by Amazon's Mechanical Turk workers. The annotators were instructed to (1) describe the image in a single sentence including the main characters, the setting or the relation of the objects; (2) If possible, include adjectives such as colors, spacing, emotion, or quantity; (3) pay attention to grammar and spelling.


\textbf{Flickr30k} \cite{young2014image} is a large English image-caption dataset. The objects in each image are segmented using bounding boxes and classified into one of 10 classes. Annotators were crowdsource workers and were asked to ``write sentences that describe the depicted scenes, situations, events and entities (people, animals, other objects)''.

Multi30k \cite{elliott2016multi30k} is a German version of Flickr30k. It contains both original and translated captions. Translations are generated by professional translators, original captions were generated by crowdworkers via the Crowdflower platform. Instructions were translated from the English instructions of Flickr30k.

Flickr8kcn \cite{li2016adding} is a Chinese version of the smaller Flickr8k dataset on which the Flickr30k dataset was based. Descriptions were generated by crowdworkers that were asked to ``write sentences describing salient objects and scenes in every image, from their own point of views''.

\textbf{MSCOCO} \cite{lin2014microsoft} is another large English image-caption dataset with additional annotations (object classes and bounding boxes). The captions were generated using human subjects on Amazon's Mechanical Turk. The annotators were given the following instructions:
\begin{itemize}
    \item Describe all the important parts of the scene.
    \item Do not start the sentences with ``There is''.
    \item Do not describe unimportant details.
    \item Do not describe things that might have happened in the future or past.
    \item Do not describe what a person might say.
    \item Do not give people proper names.
    \item The sentences should contain at least 8 words.
\end{itemize}

COCO-CN \cite{li2019coco} is a Chinese version of MSCOCO, annotated by a group of volunteers and paid undergraduate students. Annotators were instructed that the caption shall cover the main objects, actions and scene in a given image, and were provided with suggested captions retrieved in the following process: all the captions in the original MSCOCO dataset were machine-translated to Chinese, and the 5 most relevant suggestions for each image were chosen by a model. However, they were asked to provide their own descriptions, and only draw inspiration from the suggestions. In addition, they manually translated 5000 captions.

YJCaptions \cite{miyazaki2016cross} is a Japanese version of MSCOCO. Captions were generated using Yahoo! crowdsourcing, where signing up requires a Japanese proficiency, leading the authors to assume that participants were fluent in Japanese. Annotation guidelines can be translated to English as ``Please explain the image using 16 or more Japanese characters. Write a single sentence as if you were writing an example sentence to be included in a textbook for learning Japanese. Describe all the important parts of the scene; do not describe unimportant details. Use correct punctuation. Write a single sentence, not multiple sentences or a phrase''.

STAIR-captions \cite{yoshikawa2017stair} is another Japanese version of MSCOCO. Annotation guidelines can be translated to English as ``(1) A caption must contain more than 15 letters. (2) A caption must follow the da/dearu style (one of the writing styles in Japanese). (3) A caption must describe only what is happening in an image and the things displayed therein. (4) A caption must be a single sentence. (5) A caption must not include emotions or opinions about the image''.


\textbf{AIC-ICC} \cite{wu2017ai} is a large Chinese image--caption dataset. The annotators were instructed to (1) include key objects/attributes, locations and human actions; (2) generate fluent captions; (3) use Chinese idioms or descriptive adjectives.

\section{Additional Visual Examples} \label{sec:app_visual_examples}

\begin{figure} [tb]
    \centering
    \includegraphics[width=7cm]{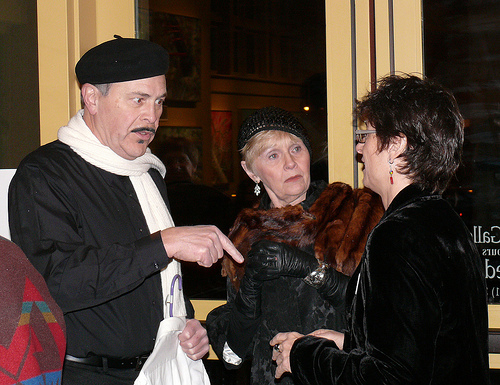}
    \caption{An image taken from Flickr30k.
    The English caption splits participants based on gender: ``A man in a beret and thin mustache gestures to two women in conversation''.
    The Chinese caption does not split participants at all:
    \begin{CJK*}{UTF8}{gbsn}
    ``三个人正在谈话''
    \end{CJK*}
    (Three people are talking).}
    \label{fig:part_dis1a}
\end{figure}

\begin{figure} [tb]
    \centering
    \includegraphics[width=5cm]{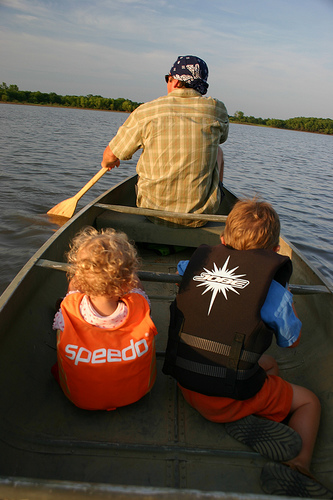}
    \caption{An image taken from Flickr30k.
    The English caption splits participants based on age: ``A man and two children in life jackets in a boat on a lake''.
    The Chinese caption does not split participants at all:
    \begin{CJK*}{UTF8}{gbsn}
    ``坐在船上出海的三个人''
    \end{CJK*}
    (Three people on a boat going out to the sea).}
    \label{fig:part_dis2}
\end{figure}

\paragraph{Numerals disagreement}
Further to the numerals disagreement analysis in Section~\ref{sec:insights}, we present examples of images that were described by captions in multiple languages with numeral value disagreement caused by differences in partition of the participants. For each of these images, the captions in one language do not partition the participants while the captions in the other is partitioning based on gender (Figure~\ref{fig:part_dis1a}), role (Figure~\ref{fig:part_dis1b}) or age (Figure~\ref{fig:part_dis2}).

\begin{figure} [!tb]
    \centering
    \includegraphics[width=7cm]{figures/data_analysis/numbers/disagreement/appendix/364213568.jpg}
    \caption{An image taken from Flickr30k.
    The English caption splits participants based on role: ``One dog is chasing another dog that is carrying something in its mouth along the beach''.
    The German caption does not split participants at all:
    ``Zwei weiß-braune Hunde, die am Strand laufe''
    (Two white and brown dogs running on the beach).}
    \label{fig:part_dis1b}
\end{figure}

\begin{figure} [tb]
    \centering
    \includegraphics[width=7cm]{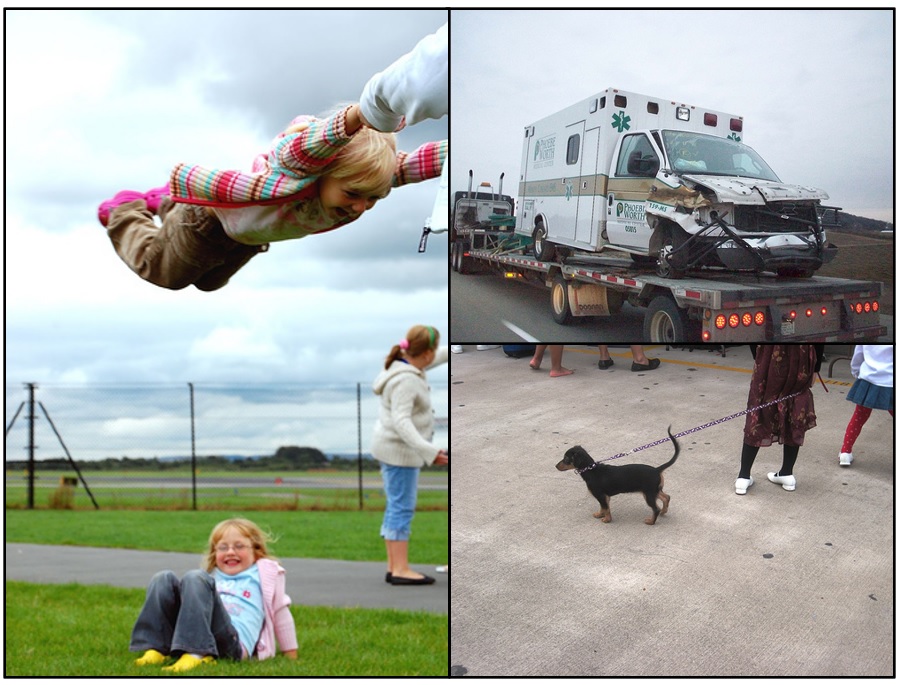}
    \caption{images with high probability for the use of passive voice. In all images, the passive participant is centered by the pose of the camera or the borders of the image.}
    \label{fig:all_passive}
\end{figure}

\paragraph{Passive voice}
Figure \ref{fig:all_passive} shows three images with high probability for the use of passive voice. In the upper right image the passive participant is centered by the pose of the camera, while in the other two images the borders of the image locates the passive participant in the center.

\section{Original vs. Translated Captions} 
\label{sec:translated_captions_analysis}

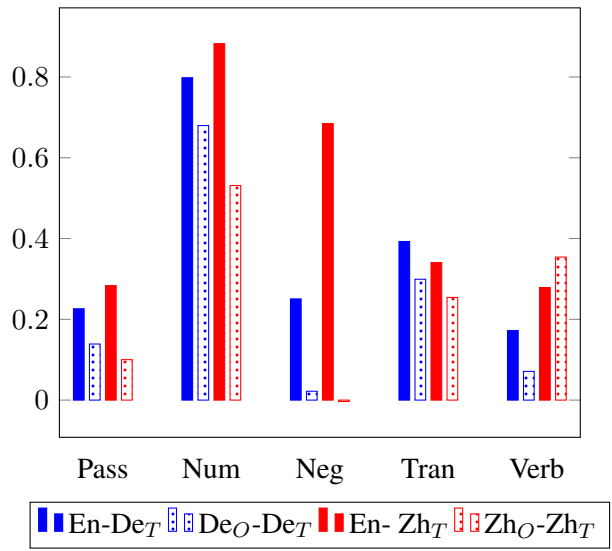
\begin{figure} [tb]
\centering
\begin{tikzpicture} 
\begin{axis} [ybar,bar width=4pt,xtick style = {draw=none},symbolic x coords = {Pass, Num, Neg, Tran, Verb}, legend style={at={(0.5,-0.15)},anchor=north,legend columns=-1}]
\addplot [draw = blue, fill = blue] coordinates {
    (Pass,0.2261) 
    (Num,0.7981)
    (Neg,0.2505) 
    (Tran,0.3927)
    (Verb,0.1721)
};
\addplot [draw = blue, pattern = dots, pattern color = blue] coordinates {
    (Pass,0.1388)
    (Num,0.6797)
    (Neg,0.0219)
    (Tran,0.2992)
    (Verb,0.0711)
};
\addplot [draw = red, fill = red] coordinates {
    (Pass,0.2834) 
    (Num,0.8825)
    (Neg,0.6846) 
    (Tran,0.3404)
    (Verb,0.2786)
};
\addplot [draw = red, pattern = dots, pattern color = red] coordinates {
    (Pass,0.1001)
    (Num,0.5312)
    (Neg,-0.0035)
    (Tran,0.2542)
    (Verb,0.3541)
};
\legend {En-De$_T$, De$_O$-De$_T$, En- Zh$_T$, Zh$_O$-Zh$_T$};
\end{axis}
\end{tikzpicture}
\caption{English -- Translated agreement (En-De$_T$ and En-Zh$_T$) and Original -- Translated agreement (De$_O$-De$_T$ and Zh$_O$-Zh$_T$) for German and Chinese, for different linguistic properties.}
\label{fig:translated_agreement}
\end{figure}

When studying multimodal tasks in non-English languages (e.g., multimodal machine translation~\cite{hitschler2016multimodal}, visual question answering~\cite{gupta2020unified}), it is common to translate an existing English image-caption corpus into the target language using crowd sourcing or translation APIs. We show that captions generated in this setting are not representative of the target language.
We use the Multi30k dataset (De) and the COCO-CN dataset (Zh), both of which
contain original as well as translated captions in the target language.
We use the statistical method described in Section~\ref{sec:corpus_analysis} in the Cross-lingual analysis paragraph to compute the agreement of English and translated captions, and compare it with the agreement of original and translated captions.
As shown in Figure~\ref{fig:translated_agreement}, in 9/10 cases the English-Translated agreement is higher than Original-Translated agreement, suggesting that translated captions are not representative of the target language. The effect is most pronounced with negation.

\end{document}